\documentclass[10pt,twocolumn,letterpaper]{article}

\usepackage{graphicx}
\usepackage{multirow}
\usepackage{booktabs}
\usepackage{xcolor}
\usepackage{amsmath}
\usepackage{algorithm}
\usepackage{algpseudocode}
\usepackage{hyperref}
\usepackage[margin=0.8in]{geometry} 

\definecolor{arxivblue}{rgb}{0.21,0.49,0.74}
\hypersetup{colorlinks=true,linkcolor=arxivblue,citecolor=arxivblue,urlcolor=arxivblue}

\usepackage{fancyhdr}
\fancypagestyle{empty}{
    \fancyhf{}
    
}
\title{Faster Multi-GPU Training with PPLL: A Pipeline Parallelism Framework Leveraging Local Learning}

\author{
\begin{tabular}{c}
Xiuyuan Guo$^{1*}$, Chengqi Xu$^{1*}$, Guinan Guo$^{3*}$, Feiyu Zhu$^{4*}$, Changpeng Cai$^{5}$, Peizhe Wang$^{5}$,\\ 
Xiaoming Wei$^{2}$, Junhao Su$^{2\dagger}$, Jialin Gao$^{2\dagger}$ \\
\end{tabular}\\
\vspace{0.5em}
\begin{tabular}{c}
$^1$University of Southern California \quad 
$^2$Meituan \quad 
$^3$Sun Yat-sen University \\
$^4$University of Shanghai for Science and Technology \quad 
$^5$Southeast University \\
\end{tabular}\\
\vspace{0.5em}
{\tt\small sujunhao02@meituan.com, gaojialin04@meituan.com} 
}

\date{} 
\begin{document}

\maketitle
\renewcommand{\thefootnote}{} 
\footnotetext{$^*$Equal contribution.}
\footnotetext{$^\dagger$Corresponding author.}
\renewcommand{\thefootnote}{\arabic{footnote}} 
\thispagestyle{empty} 

\begin{abstract}
Currently, training large-scale deep learning models is typically achieved through parallel training across multiple GPUs. However, due to the inherent communication overhead and synchronization delays in traditional model parallelism methods, seamless parallel training cannot be achieved, which, to some extent, affects overall training efficiency. To address this issue, we present PPLL (Pipeline Parallelism based on Local Learning), a novel framework that leverages local learning algorithms to enable effective parallel training across multiple GPUs. PPLL divides the model into several distinct blocks, each allocated to a separate GPU. By utilizing queues to manage data transfers between GPUs, PPLL ensures seamless cross-GPU communication, allowing multiple blocks to execute forward and backward passes in a pipelined manner. This design minimizes idle times and prevents bottlenecks typically caused by sequential gradient updates, thereby accelerating the overall training process. We validate PPLL through extensive experiments using ResNet and Vision Transformer (ViT) architectures on CIFAR-10, SVHN, and STL-10 datasets. Our results demonstrate that PPLL significantly enhances the training speed of the local learning method while achieving comparable or even superior training speed to traditional pipeline parallelism (PP) without sacrificing model performance. In a 4-GPU training setup, PPLL accelerated local learning training on ViT and ResNet by 162\% and 33\%, respectively, achieving 1.25x and 0.85x the speed of traditional pipeline parallelism.
\end{abstract}   
\section{Introduction}

\begin{figure}[h]
    \centering
    \includegraphics[width=0.45\textwidth]{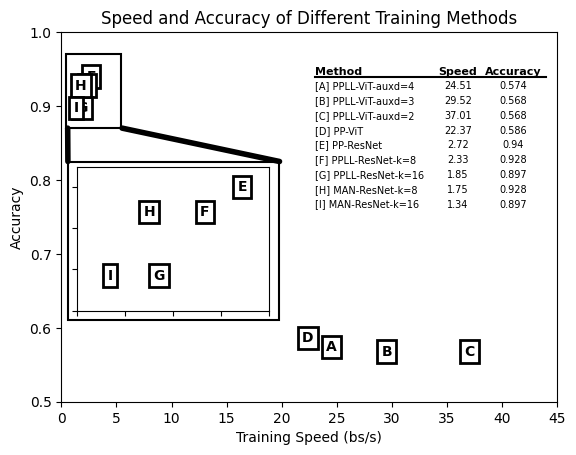}
    \caption{Comparison of training speed and test accuracy for Vision Transformer (ViT) and ResNet models using various training methods on the CIFAR-10 dataset, with a batch size of 128. This plot demonstrates that our proposed method, Pipeline parallelism based on Local Learning (PPLL), achieves significantly higher training speed with minimal accuracy trade-off, highlighting its efficiency for model training. }
    \label{head-fig}
\end{figure}

Training large-scale deep learning models efficiently across multiple GPUs is essential for advancing the capabilities of modern artificial intelligence applications. However, traditional training methods, particularly those based on Backpropagation (BP), encounter significant challenges as model sizes continue to grow. BP remains the cornerstone of training deep neural networks. Despite its effectiveness, the end-to-end (E2E) training approach presents several pressing issues. The sequential nature of BP necessitates that parameter updates for hidden layers can only commence after completing the forward and backward passes for the entire network. This sequential dependency not only hinders true parallelism but also results in substantial GPU memory consumption, as multiple layers must be stored simultaneously during training.

To address these challenges, local learning algorithms have been proposed as an alternative to the E2E approach. Local learning segments the network into discrete, gradient-isolated blocks, each updated independently by its own auxiliary network \cite{belilovsky2020decoupled}. This method enhances GPU memory efficiency and facilitates effective model parallelization by allowing different blocks to operate on separate GPUs. Current local learning methods have demonstrated promising results in image classification tasks, achieving performance comparable to E2E training \cite{hinton2006fast}. However, local learning requires auxiliary networks to maintain model performance, it also incurs additional time overhead due to the forward and backward passes of these networks, resulting in a reduction in training speed compared to E2E.

In this paper, we introduce PPLL (Pipeline
Parallelism based on Local Learning), a novel model parallel framework designed to accelerate local learning training across multiple GPUs. PPLL leverages local learning algorithms to divide the model into several distinct blocks, each assigned to a separate GPU. By implementing \textbf{ Message Queue} for data transfers between GPUs, PPLL enables seamless cross-GPU communication, allowing multiple blocks to execute forward and backward passes in a pipelined manner. This pipelined execution minimizes idle times and prevents bottlenecks typically caused by sequential gradient updates, thereby significantly accelerating the overall training process.

We validate the effectiveness of PPLL through extensive experiments using ResNet and Vision Transformer (ViT) architectures on  CIFAR-10\cite{krizhevsky2009learning},
SVHN\cite{netzer2011reading}, and STL-10\cite{coates2011analysis} datasets. Our results demonstrate that PPLL achieves performance comparable to end-to-end training while significantly enhancing training speed. These empirical findings confirm that PPLL not only reduces training time but also maintains comparable model accuracy, offering a scalable and efficient solution for distributed local learning.

The main contributions of this paper are as follows:

\textbf{Pipeline Parallelism for Accelerated local learning}: We propose a pipeline parallel training framework for the local learning method (PPLL). Leveraging the relative independence between modules in local learning, we can effectively segments the model into gradient-isolated blocks to enable parallel training across modules, thereby significantly reducing training time.
This design incorporates scalability and versatility, allowing seamless adaptation to various model architectures and enabling efficient cross-GPU communication through the use of queues.

\textbf{Theoretical Training Time Analysis}: We provide a comprehensive theoretical analysis of the training time of PPLL compared to end-to-end (E2E) training and traditional pipeline parallelism (PP). By deriving time complexity equations for processing a single batch under different training paradigms, we demonstrate that PPLL can achieve significant reductions in training time. Our analysis shows that as the number of network segments increases, the training time under PPLL decreases and can become shorter than both E2E and PP methods once a critical threshold is reached

\textbf{Empirical Validation}: We conduct comprehensive experiments on widely-used architectures and datasets, demonstrating that PPLL can greatly accelerate local learning and even faster than naive parallel training methods while maintaining high accuracy. Particularly, as the number of GPUs available for training increases, the performance gap between PPLL and PP widens further. In our  ViT experiments, when using 4 GPUs, PPLL's training speed surpasses that of PP.

Our proposed PPLL framework advances the state-of-the-art in multi-GPU local learning training methodologies, offering a robust and scalable solution for accelerating local learning model training. This work lays the groundwork for future developments in distributed local learning training techniques, enabling the efficient deployment of models across diverse computational environments.

\section{Related Work}
\label{gen_inst}
\textbf{Pipeline Parallelism.}
Complementarily, Huang et al. (2019) \cite{huang2019} propose GPipe, a library that implements pipeline parallelism to manage the training of vast neural networks that exceed the memory capacities of individual hardware accelerators. By splitting the model into different segments processed on separate GPUs and introducing a novel batch-splitting algorithm, GPipe enables the linear scaling of training processes across multiple GPUs. This method effectively reduces computational bottlenecks by balancing the load across the network and minimizing idle times through synchronized gradient updates across split batches.

\textbf{Data Parallelism.}
Data parallelism is a prevalent approach in distributed computing that accelerates model training by partitioning large datasets across multiple processors or machines. As early as several decades ago, Hillis et al. (1986) \cite{hillis1986} proposed data parallel programming, and the success of parallel algorithms demonstrated that this programming paradigm had much broader applicability than previously conceived. Krizhevsky et al. (2012) \cite{krizhevsky2012} introduced the successful application of data parallelism on the large-scale ImageNet dataset, establishing the foundation for subsequent data parallel methodologies. Li et al. (2020) \cite{li2020} designed the distributed data parallel module in PyTorch v1.5, achieving exceptional performance in evaluating NCCL and Gloo backends on ResNet50 and BERT models. The Meta AI team introduced Fully Sharded Data Parallel (FSDP) \cite{zhao2023} in PyTorch, demonstrating outstanding performance with near-linear TFLOPS scalability to support larger models. However, Data Parallelism requires each device to maintain a complete model replica, which consumes substantial memory resources for large-scale deep learning models.

\textbf{Local Learning.}
Local Learning focuses on constructing customized models for local data subsets rather than a single global model for the entire dataset. Local learning algorithms partition data into smaller regions based on spatial proximity, feature similarity, or other criteria, and then train separate models for each region \cite{illing2021, xiong2020}. Cheng et al. (2019) \cite{cheng2019} proposed Local-to-Global Learning (LGL), which fundamentally progresses from learning on fewer local samples to more global samples to enhance performance in classification tasks. Guo et al. (2023) \cite{guo2023} introduced the FedBR algorithm, which improves federated learning performance on heterogeneous data by reducing local learning bias. Patel et al. (2023) \cite{patel2023} implemented local learning by incorporating the concept of width modularization into deep modularization, thereby enhancing the efficiency of model parallel training. However, when these networks are unrestrainedly partitioned into numerous local blocks, network performance deteriorates as backpropagation fails to effectively concatenate parameters between these local blocks.

\section{Method}
\label{headings}

\subsection{Preliminaries}

In a naive pipeline parallel training process, a large network is divided into $S$ segments, denoted as \{$M_1, M_2, \cdots, M_S$\}, with each segment assigned to a separate GPU. Initially, data is fed into $M_1$, and upon completion, the results are passed to $M_2$, continuing this way until $M_S$ computes the final output. However, backward propagation only begins after all computations are completed. These $S$ stages must be executed sequentially, which results in low parallelization efficiency and underutilization of computational resources, as later blocks must wait for preceding blocks to finish before they can receive features.

In conventional local learning, the network is similarly divided into $S$ modules, and each segment can be independently updated with the assistance of a corresponding auxiliary network. Consequently, each module is relatively isolated from the others. Despite this independence, traditional local learning does not achieve parallelism.

\begin{figure}[h]
    \centering
    \includegraphics[width=0.45\textwidth]{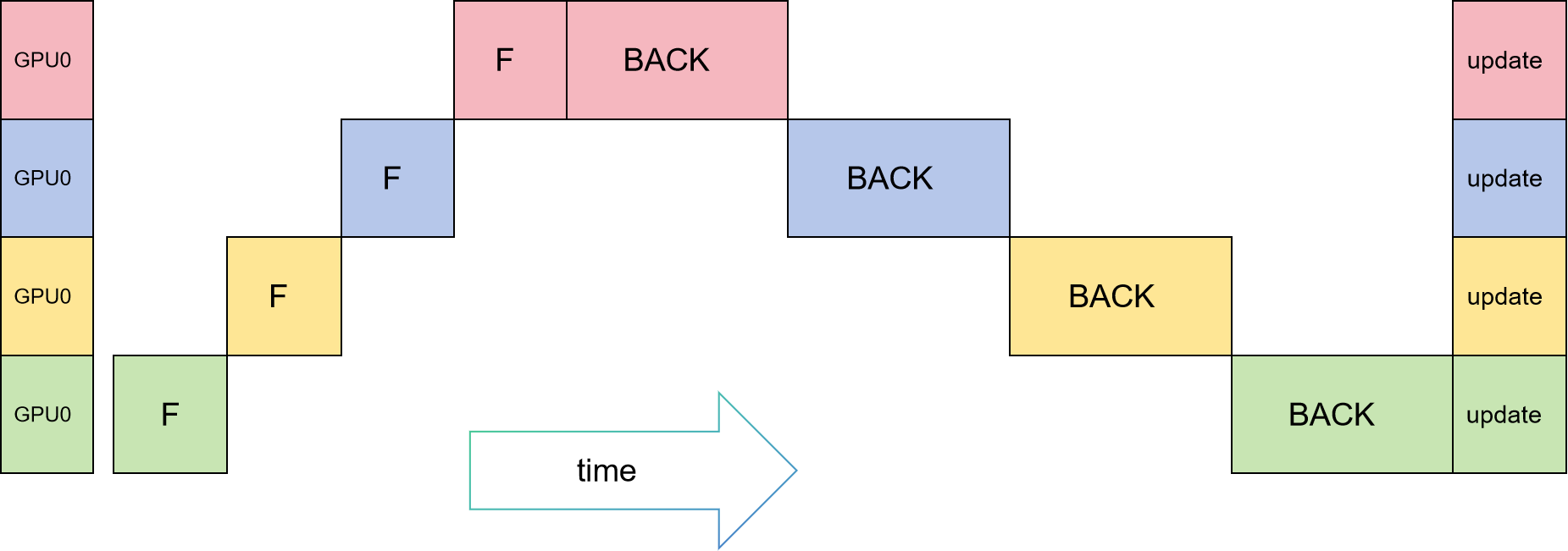}
    \caption{Naive Pipeline Parallelism}
    \label{fig:enter-label}
\end{figure}

\subsection{Pipeline Parallelism based on Local Learning (PPLL)}

To address the limitations of computational efficiency and time in the aforementioned methods, we propose a pipeline parallelism method based on local learning (PPLL). This approach builds on the local learning framework, leveraging the advantage of the independence of each module, allowing each module to update its parameters independently. In PPLL, backpropagation can commence immediately after the forward pass of each module, without waiting for the results of subsequent modules. By caching each module's output on separate GPUs, we achieve pipeline parallelism between modules, ensuring that each module can immediately access the latest or second-latest outputs without waiting for prior module calculations to complete. This design eliminates idle time during cross-GPU computations, enabling each GPU module to reach optimal computational efficiency and significantly reducing training time.

\textbf{Training Procedure.} In pipeline parallel training, each stage represents the training of a module. We divide the network into \( s \) modules, denoted as 
\{$M_1,\cdots,M_S$\}, with each modules assigned to  a separate GPU, resulting in a total of \( s \) stages. Buffers \{$B_1,\cdots,B_{S}$\} are introduced between modules to store outputs, allowing modules to begin computations independently. At each iteration,  module $M_j$ gets input from the buffer $B_{j-1}$ and performs a forward pass, store the output $x_{j}$ in buffer  $B_{j}$, then computes local losses, and updates its parameters immediately without waiting for other modules. This enables simultaneous forward and backward computations across different modules, maximizing computational efficiency. The detailed steps are provided in the following pseudo-code.

\begin{algorithm}
\caption{Streaming Algorithm with Pipeline Parallelism based on Local Learning}
\begin{algorithmic}[1]
\Require Stream $S \{(x_0^t, y^t)\}_{t \leq T}$; Buffer size $M$; modules $M_{j \leq S}$
\State Initialize: Buffers $\{B_j\}_{j}$; params$\{\theta_j, \gamma_j\}_{j}$; $P_j \leftarrow M_j $
\State \textbf{Parallel Execution:} $\forall j \in \{1, \dots, K\}, P_j$
\While{training}
    
    \If{$j = 1$}
        \State $(x_0, y) \gets S$ \Comment{Get input data from the stream for the first module}
    \Else
        \State $(x_{j-1}, y) \gets B_{j-1}$ \Comment{Get input from the previous module's buffer}
    \EndIf
    \State $x_j \gets M_{j}(x_{j-1})$ \Comment{Perform forward pass for module $M_{j}$}
    
    \If{$j < K$}
        \State $B_j \gets (x_j, y)$ \Comment{Store the output in buffer $B_j$ for the next module}
    
    \EndIf
    \State Compute $\nabla_{(\gamma_j, \theta_j)} \hat{\mathcal{L}}^t(y, x_j; \gamma_j, \theta_j)$ \Comment{Perform backward and calculate gradient}
    \State $(\theta_j, \gamma_j) \gets$ Update parameters $(\theta_j, \gamma_j)$ \Comment{Update parameters for module $M_j$}
\EndWhile
\end{algorithmic}
\end{algorithm}

\textbf{Achieving Fully Pipelined Training via Local Learning}. Traditional pipeline parallelism methods suffer from idle time due to the sequential dependency of forward and backward passes across modules, particularly when scaling across multiple GPUs. 

In our proposed PPLL approach, we decouple the backward computations of each module by introducing local learning, where each module computes its own local loss and updates its parameters independently. By caching the outputs between modules, we eliminate the need for modules to wait for the completion of previous modules' backward passes. Each module can immediately start its backward pass after its forward computation, using the most recent outputs available. This design effectively transforms the training process into a fully pipelined system, where forward and backward computations are overlapped across modules and GPUs

\begin{figure}[h]
    \centering
    \includegraphics[width=0.48\textwidth]{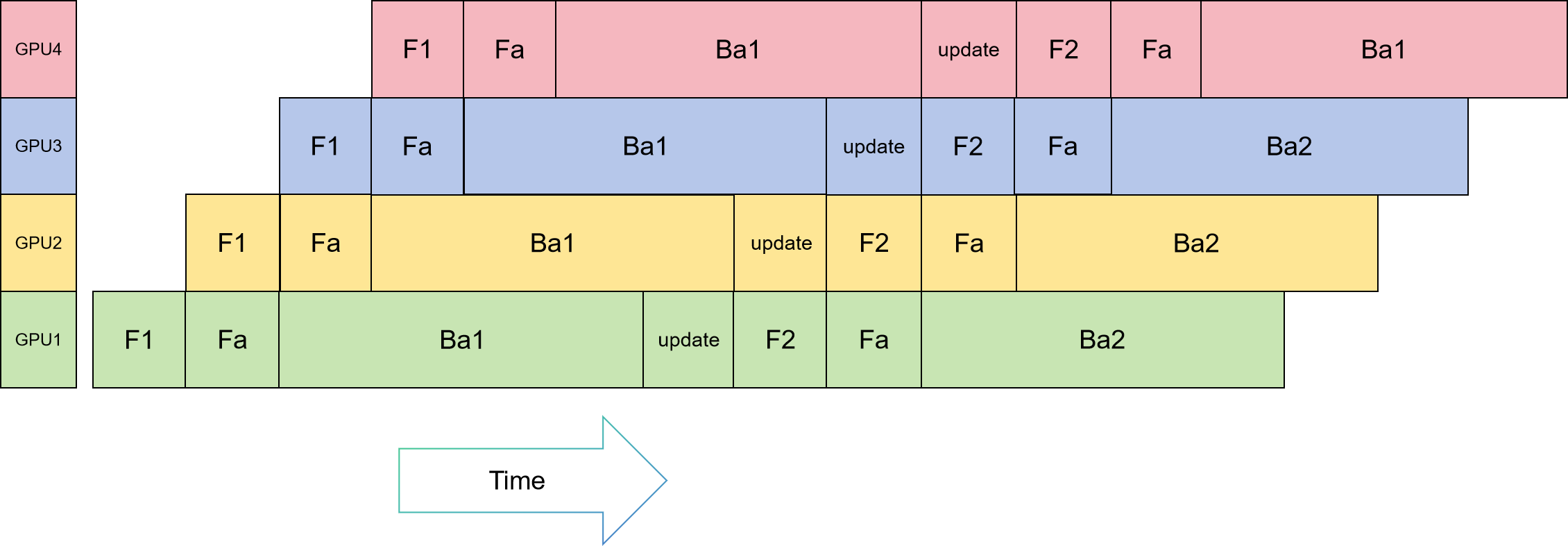}
    \caption{PIPELINE PARALLELISM BASED ON LOCAL LEARNING}
    \label{fig:enter-label}
\end{figure}

\subsection{Theoretical Training Time Analysis}

To compare the efficiency of PPLL with end-to-end (E2E) training and traditional pipeline parallelism, we analyze the time complexity for processing a single batch.

Let \( f_i \) and \( b_i \) denote the forward and backward pass times for module \( M_i \), respectively.

Assume an optimal partitioning where each module has approximately equal computational complexity (closely proportional to the number of trainable parameters). In both pipeline parallelism (PP) and PPLL, each module \( M_i \) is assigned to one of \( s \) stages, with \( s \) distinct stages in total.

In PPLL, each stage also includes an auxiliary network \( A_i \) to assist in updating \( M_i \).

\textbf{End-to-End Training}:
In E2E training, all \( s \) modules complete both forward, backward, and parameter update operations sequentially. Thus, the total time to process one batch is:

\begin{equation}
S_{\text{E2E}} = F + B + U = \sum_{i=1}^{s} f_i + B + U
\end{equation}

where \( F \) is the time cost of the forward pass, \( B \) is the time cost of the backward pass, and \( U \) is the time cost of the parameter update.

\textbf{Naive Pipeline Parallelism}:
In naive pipeline parallelism, the \( s \) modules are assigned to \( s \) distinct devices, with each module executing sequentially, and intermediate activations passing between devices with a communication overhead \( Q(s) \). The total time to process one batch is:

\begin{equation}
S_{\text{PP}} = \sum_{i=1}^{s} f_i + B + U + Q(s)
\end{equation}

\textbf{PPLL}:
PPLL leverages the independence of backward passes, reducing idle time by allowing the backward pass of a module to begin before the subsequent module’s backward pass completes. This forms a pipelined execution between stages. The training time for one batch is therefore equal to the execution time of the first stage (which is typically the longest). The total time for one batch is:

\begin{align}
S_{\text{PPLL}} &= F_{a1} + B_{a1} + U_{a1} + Q(s) \notag \\
                &= f_1 + f_{a1} + B_{a1} + U_{a1} + Q(s)
\end{align}

where \( f_1 \) and \( f_{a1} \) represent the forward times of the module in stage 1 and its auxiliary network, respectively, \( B_{a1} \) denotes the backward time for both the module and auxiliary network in stage 1, and \( U_{a1} \) represents the parameter update time for both.

\textbf{Comparison of PPLL and PP}:

Let \( f_{a1} = A_{1}(s) \) denote the forward time of the auxiliary network associated with first module \( M_1 \). As the number of stages \( s \) increases, the number of layers in first module \( M_1 \) decreases, reducing the size of the corresponding auxiliary network. Therefore, \( A_{1}(s) \) is a decreasing function of \( s \).

Let \(F_{a1}\) denote the forward time of the combination of \( M_1 \) and its auxiliary network.

The time difference between PP and PPLL is given by:

\begin{align}
S_{\text{PPLL}} &= F_{a1} + B_{a1} + U_{a1} + Q(s) \notag \\
                &= f_1 + f_{a1} + B_{a1} + U_{a1} + Q(s)
\end{align}

If 

\[
(F - F_{a1}) > 0 \Rightarrow \frac{s-1}{s}F > A_{1}(s),
\]

then:

\begin{equation}
B > B_{a1}\Rightarrow B - B_{a1} > 0 
\end{equation}
\begin{equation}
U > U_{a1}\Rightarrow U - U_{a1} > 0 
\end{equation}
\begin{equation}
F > f_{1} + f_{a1} \Rightarrow \sum_{i=2}^{s} f_i - f_{a1} > 0 
\end{equation}

This yields:

\begin{equation}
T_{\text{PPLL}} < T_{\text{PP}}
\end{equation}

Thus, as the number of stages \( s \) increases, the training time under the PPLL framework progressively decreases and eventually becomes shorter than that of both the E2E and PP methods once the critical threshold is reached.

\subsubsection{Simplified Training Time Analysis}

To quantify the performance gain of PPLL over naive Pipeline Parallelism (PP) under ideal conditions, we introduce the following assumptions and simplifications:

\begin{itemize}
    \item (1) We assume that after dividing the network into $k$ segments, the sum of the backward pass times for each segment, denoted as $B_i$, equals the total backward pass time for the entire network, i.e., $B = \sum_{i=0}^{k} B_i$.
    
    \item (2) Similarly, we assume that the sum of the parameter update times for each segment, denoted as $U_i$, equals the total parameter update time for the entire network, i.e., $U = \sum_{i=0}^{k} U_i$.
    
    \item (3) We assume that the computational complexity of the auxiliary network $a$ for the first module $M_1$ is $k$ times that of $M_1$. We further assume this linear relationship applies to the forward pass, backward pass, and parameter update computations, i.e., $f_{a} = kf_{1}$, $b_{a} = kb_{1}$, and $u_{a} = ku_{1}$.
    
    \item \textbf{This analysis ignores communication costs}, such as the time required to transfer tensors across devices and the time needed to store and retrieve data from buffers.
\end{itemize}

Using these assumptions, we analyze the time ratio of PPLL to PP:

\begin{align}
    \frac{T_{\text{PPLL}}}{T_{\text{PP}}} &= \frac{ f_{1} + f_{a} + B_{a1} + U_{a1} }{F + B + U} \\
    &= \frac{ f_{1} + f_{a} + b_{a} + b_{1} + u_{a} + u_{1} }{F + B + U} \notag\\
    &\quad (\text{by assumptions (1) and (2)}) \\
    &= \frac{ f_{1} + kf_{1} + kb_{1} + b_{1} + ku_{1} + u_{1} }{sf_{1} + sb_{1} + su_{1}} \notag\\
    &\quad (\text{by assumption (3)}) \\
    &= \frac{k+1}{s}.
\end{align}

This result indicates that, under ideal conditions, the training time of the PPLL framework is $(k+1)/s$ times that of PP. Thus, as the number of network segments increases (i.e., $s$ becomes larger) and the size of the auxiliary network becomes smaller, the ratio decreases, reflecting improved efficiency.

\textbf{Note}: This simplified analysis assumes ideal conditions and ignores communication costs associated with data transfer and synchronization between modules. In practical implementations, these overheads can affect the actual performance gains. Therefore, empirical evaluations are necessary to assess the real-world effectiveness of PPLL.

\section{Experiment}
\label{others}

\subsection{Experimental Setup}
We perform experiments on three commonly used datasets: CIFAR-10 [21], SVHN [30], and STL-10 [9], using ResNet-32 [15] and ViT-base-patch16-224 models with varying depths. we employ the SGD optimizer with Nesterov momentum set at 0.9 and an L2 weight decay of 1e-4. For the experiement of Resnet, the batch sizes are 1024 for CIFAR-10 and SVHN, and 128 for STL-10. For ViT, the batch sizes of all datasets are 128. Training lasts for 400 epochs, with initial learning rates set to 0.01, using a cosine annealing learning rate schedule [9]. To compare our pipeline parallelism, we evaluate the same configurations against standard supervised local learning methods, naive pipeline parallelism, and E2E training, ensuring consistent training settings to avoid any confounding factors.

\subsection{Implement Details}
For the ResNet experiments, we use MAN (Momentum Auxiliary Network for Supervised Local Learning) \cite{su2024momentum} as the baseline model, conducting experiments with both 8-local-module and 16-local-module versions of MAN. For the ViT experiments, we implemented our own local learning version of ViT with 12 local blocks. For the auxiliary network of ViT, we use N attention blocks followed by a classifier. We define N here as the depth of the auxiliary network. We come up with a heuristic way to calculate N for each layer l: 

\begin{equation}
    N_l = d' - \lfloor l / n \rfloor
\end{equation}

where l represents the current layer, n is a hyperparameter, here we use 3 and d' is the max depth of the auxiliary network. This formula is inspired by the idea that the former blocks need more information. In our experiment, we evaluated the performance at maximum auxiliary network depths of 2, 3, and 4.

In our ViT experiments, we implemented the largest Pipeline Parallelism(LPP), meaning that each local block was assigned to a separate process, enabling parallel training of each local block under specific GPU resources.

\subsection{Evaluation Metrics}
To thoroughly assess the performance of our model acceleration techniques, we employ three key evaluation metrics:

\textbf{Batches per second (bs/s)}: This metric measures the number of batches processed per second, providing a direct indication of the training speed and efficiency of our parallelism strategies.

\textbf{Memory utilization}: We monitor the maximum memory consumption during training of each GPU, as efficient memory usage is crucial for large-scale models and datasets, particularly when dealing with deep architectures like ResNet and ViT.

\textbf{Accuracy}: To ensure that our acceleration techniques do not compromise model performance, we track the final accuracy of the model on the test sets of the CIFAR-10, SVHN, and STL-10 datasets. Accuracy is a standard metric to evaluate the effectiveness of the trained model.

By analyzing these metrics, we can comprehensively evaluate both the speed-up and the resource efficiency of our proposed parallelism methods, while ensuring the model's predictive performance remains robust.

\subsection{Results on Image Classification Task}

Our experimental results demonstrate that PPLL achieves significant training speedups across all datasets (CIFAR-10, SVHN, STL-10, and ImageNet) and model architectures (ResNet-32 and ViT), particularly when scaled to multi-GPU setups. As shown in Table \ref{spd-table}, PPLL achieves a training speed improvement over MAN of up to 33\% on CIFAR-10 with 4 GPUs. Similarly, in ViT experiments on CIFAR-10 with 4 GPUs, PPLL achieves a notable 162\% speedup compared to local learning method S(d=4), even 26.0\% surpass naive pipeline parallelism and outperforms PP as the model depth and auxiliary network depth increase.

In terms of Accuracy, PPLL exhibits comparable performance gains. PPLL achieves validation accuracy comparable to the E2E method under nearly all auxiliary network configurations. These results underscore the effectiveness of PPLL in maximizing training efficiency without sacrificing model performance, even in resource-constrained environments.

Furthermore, in the context of memory utilization, PPLL demonstrates approximately a 10\% reduction in GPU memory usage per GPU compared to standard local learning. This reduction in memory overhead allows for more efficient parallel processing and greater scalability in training larger models on limited GPU resources.

Overall, these results validate the efficacy of PPLL as a scalable, high-performance training framework, capable of delivering both speed and accuracy in multi-GPU settings.

\begin{table*}
\centering
\small 
\renewcommand{\arraystretch}{1.2} 
\begin{tabular}{|c|c|c|c|c|c|c|c|}
\hline
Dataset & Stage & \multicolumn{3}{c|}{ResNet-32} & \multicolumn{3}{c|}{ViT} \\ \hline
 &  & PP & PPLL ($k=8$) & PPLL ($k=16$) & PP & ViT ($d=4$) & ViT ($d=2$) \\ \hline
CIFAR-10 & E2E ($S=1$)  & 2.83 bs/s & 1.75 bs/s & 1.34 bs/s & 23.3 bs/s & 9.54 bs/s & 16.81 bs/s \\ \hline
 & 2 GPUs ($S=2$) & 2.74 bs/s & 2.22 bs/s & 1.70 bs/s & 25.5 bs/s & 12.76 bs/s & 22.30 bs/s \\ \hline
 & 4 GPUs ($S=4$) & 2.72 bs/s & \textbf{2.33} bs/s & \textbf{ 1.85} bs/s & 20 bs/s & \textbf{22.32} bs/s & \textbf{32.12} bs/s \\ \hline
 & 4 GPUs (LPP) & - & - & - & - & \textbf{25.21} bs/s & \textbf{37.60} bs/s \\ \hline
SVHN & E2E ($S=1$)  & 2.87 bs/s & 1.80 bs/s & 1.26 bs/s  & 24.40 bs/s & 7.35 bs/s & 13.47 bs/s \\ \hline
 & 2 GPUs ($S=2$) & 2.85 bs/s & 2.24 bs/s & 1.75 bs/s & 24.00 bs/s & 13.28 bs/s &24.01 bs/s \\ \hline
 & 4 GPUs ($S=4$) & 2.85 bs/s & \textbf{2.35} bs/s & \textbf{ 1.82} bs/s & 20.50 bs/s & \textbf{28.43 }bs/s &\textbf{ 38.39} bs/s \\ \hline
 & 4 GPUs (LPP) & - & - & - & - &\textbf{ 31.43} bs/s & \textbf{40.54} bs/s \\ \hline
STL-10 & E2E ($S=1$)  & 5.00 bs/s & 2.66 bs/s & 1.81 bs/s & 9.65 bs/s & 2.21 bs/s & 4.00 bs/s \\ \hline
 & 2 GPUs ($S=2$) & 4.86 bs/s & 4.00 bs/s & 2.10 bs/s & 9.60 bs/s & 4.44 bs/s & 7.22 bs/s \\ \hline
 & 4 GPUs ($S=4$) & 5.00 bs/s & \textbf{4.24} bs/s &  \textbf{2.23} bs/s & 9.60 bs/s & 8.37 bs/s & \textbf{10.01} bs/s \\ \hline
 & 4 GPUs (LPP) & - & - & - & - & 6.50 bs/s & \textbf{10.55} bs/s \\ \hline
\end{tabular}
\caption{Comparison of Training Speeds for Different Configurations of ResNet-32 and ViT on CIFAR-10 and STL Datasets. For the ViT part, ViT($d=4$) means the maximum depth of the auxiliary network is 4.}
\label{spd-table}
\end{table*}

\subsubsection{Training-Speed Analysis}
We start by assessing the training speed of our approach using the CIFAR-10 \cite{krizhevsky2009learning}, SVHN\cite{netzer2011reading}, and STL-10 \cite{coates2011analysis} datasets. We employ ResNet-32 \cite{he2016deep}, partitioned into 8 and 16 local blocks, and ViT(ViT-Base model, patch size of 16 and an input resolution of 224×224), with 3 to 4 maximum auxiliary network depth . As illustrated in Table \ref{spd-table}, we provide the following analysis:

On the CIFAR-10 dataset, our method demonstrates significant improvements in training speed across various configurations, particularly when scaling the number of GPUs. For ResNet-32 partitioned into 8 local blocks ($k=8$), our method achieves a 26.8\% and 33.1\% increase in training speed compared to the original local learning network when using 2 GPUs and 4 GPUs, respectively, reaching 84\% of the speed of PP. For the ViT model with a depth of $d=4$, our method shows notable improvements as well. On 4 GPUs ($S=4$) with a maximum auxiliary depth of 4, the training speed increases from $9.54~\text{bs/s}$ (PP) to \textbf{22.32~\text{bs/s}}, achieving a \textbf{134\%} enhancement. When employing layer-wise pipeline parallelism (LPP) on 4 GPUs, we reach the highest training speeds of \textbf{25~\text{bs/s}} and \textbf{37~\text{bs/s}} respectively with maximum auxiliary depths of 4 and 2. A similar trend is observed on the STL-10 dataset. For the ViT model with a maximum auxiliary depth of 2, the training speed on 4 GPUs ($S=4$) increases from $9.60~\text{bs/s}$ (PP) to \textbf{10.01~\text{bs/s}}, achieving a modest improvement. When utilizing Largest pipeline parallelism (LPP) on 4 GPUs, the training speed further enhances to \textbf{10.55~\text{bs/s}}, surpassing the PP method.

Consistent with the CIFAR-10 results, we observe similar findings on the SVHN and STL-10 datasets. For ResNet, we observe a highest increase of 30.5\% on SVHN and 59.3\% on STL-10 with 4 GPUs and local module $k=8$. As for ViT, we achieve the highest improvements of 200.9\% and 163\% on these two datasets with a maximum auxiliary depth of 2. Notably, the speed of PPLL is nearly twice that of PP on SVHN.

Notably, from a theoretical perspective, dividing a network into two or four parts with equal computational complexity should result in PPLL speeds approximately 2 or 4 times that of the original network. However, our experiments revealed that when using message queues for inter-process communication, intermediate variables need to be transferred from the GPU to the CPU before being stored in the cache. The time complexity of this transfer is positively correlated with the size of the intermediate variables. Therefore, in cases where the model is relatively small but the intermediate variables are large (such as ResNet-32, where the size of intermediate variables is $1024 \times 32 \times 16 \times 16$, which is relatively large), the communication overhead accounts for a non-negligible proportion of the training time per batch. This explains why the speed improvement in the ResNet experiments is not significant.

In contrast, in the ViT experiments, because the intermediate variables are smaller ($128 \times 5 \times 768$), the communication overhead is relatively minor, resulting in a more significant speed improvement. In the ViT experimental results, we can verify the previously derived formula for the time ratio of PPLL to PP. When $S=2$, $d=2$, and $k=2$, theoretically, $\frac{\text{PPLL}}{\text{PP}} = \frac{1}{2}$, which is close to the actual value $\frac{12.76}{25.5}$. Similarly, when $S=2$, $d=2$, and $k=2$, theoretically, $\frac{\text{PPLL}}{\text{PP}} = 1$, which aligns with the actual value $\frac{22.30}{25.5}$.

These results highlight the effectiveness of our method in accelerating training speeds for ResNet and ViT models, particularly in multi-GPU environments.

\subsubsection{Training Accuracy Analysis}

Based on the experimental results presented in Table \ref{vit-acc} and Table \ref{vit-acc2}, our pipeline parallel learning (PPLL) method achieves comparable accuracy to the end-to-end (E2E) approach across different configurations and datasets.

For the CIFAR-10 dataset, PPLL with different ViT configurations (depths $d=2,3,$ and $4$) maintains training and testing accuracies similar to those of the E2E approach, with only minimal reductions. Specifically, PPLL-ViT (with $d=4$) achieves a test accuracy of $0.574$, very close to the $0.578$ achieved by E2E, demonstrating that PPLL can deliver nearly equivalent performance while offering more flexible parallelization options.

Similarly, on the SVHN dataset, PPLL shows comparable test accuracies to the E2E approach, with slight variations across configurations. For example, PPLL-ViT (with $d=4$) achieves a test accuracy of $0.847$ versus $0.861$ in E2E, confirming that PPLL’s accuracy is on par with E2E even as it enables effective multi-GPU utilization.

In the case of ResNet-32 configurations, PPLL achieves training and testing accuracies that are comparable to E2E on the CIFAR-10 dataset. For instance, PPLL-ViT (with $k=8$) records a test accuracy of $0.86$, closely matching the $0.91$ of E2E, thus demonstrating that our PPLL method is effective in maintaining high performance across different model architectures.

In summary, our PPLL method is capable of achieving accuracy levels comparable to the traditional E2E training method across various configurations and datasets. This makes PPLL a valuable approach for distributed training, as it combines accuracy retention with the benefits of improved parallelization and scalability.

\begin{table}[H]
\centering
\setlength{\tabcolsep}{1.5pt} 
\small 
\begin{tabular}{cccc}
\hline
Dataset & Method & Train Accuracy & Test Accuracy \\ 
\hline
\multirow{4}{*}{CIFAR-10}
& PP (E2E) & 1.000 & 0.578 \\
& PPLL-ViT(d = 4) & 0.960 & 0.574 \\
& PPLL-ViT(d = 3) & 0.958 & 0.568 \\
& PPLL-ViT(d = 2) & 0.958 & 0.568 \\
\hline
\multirow{4}{*}{SVHN}
& PP (E2E) & 1.000 & 0.861 \\
& PPLL-ViT(d = 4) & 0.982 & 0.847 \\
& PPLL-ViT(d = 3) & 0.982 & 0.844 \\
& PPLL-ViT(d = 2) & 0.982 & 0.843 \\
\hline
\multirow{3}{*}{STL-10}
& PP (E2E) & 1.000 & 0.416 \\
& PPLL-ViT(d = 4) & 0.895 & 0.401\\
& PPLL-ViT(d = 3) & 0.913 & 0.414 \\
& PPLL-ViT(d = 2) & 0.909 & 0.413 \\
\hline
\end{tabular}
\caption{Comparison of Train Accuracy and Test Accuracy for Different Configurations of ViT on CIFAR-10 and SVHN Datasets. PPLL-ViT(d=4) means the maximum auxiliary network depth is 4. }
\label{vit-acc}
\end{table}

\subsubsection{GPU Memory Requirement Analysis}
Supervised local learning limits gradient propagation to within local blocks, which shortens the path of backpropagation, and thereby markedly conserving GPU memory. Our proposed method further conserves GPU memory usage on a single GPU by dividing the model into blocks for parallel training. Our analysis, encompassing a thorough comparison of end-to-end training, pipeline parallel training, and PPLL training on the CIFAR-10 dataset, demonstrates that PPLL implementation effectively reduces memory consumption on a single GPU. In conjunction with the analysis presented in previous sections, PPLL not only accelerates training but also alleviates the memory burden on a single GPU, with only a minimal trade-off in model performance.\\

\begin{table}[H]
\centering
\setlength{\tabcolsep}{0pt} 
\small 
\begin{tabular}{cccc}
\hline
Dataset & Method & Train Accuracy & Test Accuracy \\ 
\hline
\multirow{3}{*}{CIFAR-10} 
& PP (E2E) & 1.000 & 0.940 \\
& PPLL-Resnet(k=8) & 0.995 & 0.928 \\
& PPLL-Resnet(k=16) & 0.965 & 0.897 \\
\hline
\multirow{3}{*}{SVHN}
& PP (E2E) & 1.000 & 0.973 \\
& PPLL-Resnet(k=8) & 1.000 & 0.971 \\
& PPLL-Resnet(k=16) & 0.982 & 0.956 \\
\hline
\multirow{3}{*}{STL-10}
& PP (E2E) & 1.000 & 0.865 \\
& PPLL-Resnet(k=8) & 1.000 & 0.850 \\
& PPLL-Resnet(k=16) & 1.000 & 0.843 \\

\hline
\end{tabular}
\caption{Comparison of Train Accuracy and Test Accurany for Different Configurations of ResNet-32 on CIFAR-10 and SVHN Datasets}
\label{vit-acc2}
\end{table}

\begin{table}[H]
    \centering
    \setlength{\tabcolsep}{2pt} 
    \small 
    \begin{tabular}{|l|c|c|c|}
        \hline
        \textbf{Method} & \textbf{ViT-Auxdp4} & \textbf{ViT-Auxdp3} & \textbf{ViT-Auxdp2}\\ 
        \hline
        \textbf{PP(s=2)} & 2293 & 2293 & 2293 \\ 
        \textbf{PP(s=4)} & 1737 & 1737 & 1737 \\ 
        \textbf{PPLL(s=2)} & 3141 & 2579 & 2099 \\
        \textbf{PPLL(s=4)} & 2121 & 1841 & 1541 \\
        \hline
    \end{tabular}
    \caption{Comparison of single GPU memory usage using ViT as a backbone on the CIFAR-10 dataset. PP represents pipeline parallelism, and PPLL represents pipeline parallelism based on Local Learning. For the ViT part, Auxdp2 means the maximum depth for the auxiliary network is 2. Because PP doesn't contain an auxiliary network, the values in its row are all the same, representing the GPU usage of PP.}
    \label{vit-gpu}
\end{table}


Table \ref{vit-gpu} displays the GPU memory consumption observed during the training of a Vision Transformer (ViT) model on the CIFAR-10 dataset, comparing the conventional pipeline parallelism (PP) with our proposed method, PPLL. In this setup, the ViT model is partitioned into 4 blocks. This partitioning strategy is adopted to avoid the high memory overhead associated with a 12-block division, where each block would necessitate maintaining separate copies of data and training-related information, including gradient and activation values. \\
\par The results demonstrate that PPLL achieves an approximate 10\% reduction in GPU memory consumption on a single GPU when the maximum depth of the auxiliary network is set to 2. However, as the maximum depth of the auxiliary network exceeds 2, memory usage increases. This rise in memory consumption occurs because increasing network depth not only expands the number of model parameters but also extends the backpropagation path. In conjunction with the findings presented in Table \ref{vit-acc}, utilizing an auxiliary network with a maximum depth of 2 incurs only a minimal reduction in accuracy. \\

       

\section{Conclusion}
This paper proposes PPLL (Pipeline Parallelism based on Local Learning), a method for more efficient multi-GPU parallel training using local learning. PPLL places the gradient-truncated partitioned network on different GPUs, where each GPU stores the latest partial features of the current network block in a queue. This setup allows each GPU to only receive features stored in the queue of the previous GPU, without the need for backpropagation between GPUs, thus enabling seamless cross-GPU communication and accelerating multi-GPU training. We validate our performance on different architectures and image classification datasets, demonstrating that we can accelerate training without sacrificing model performance.

\noindent {\bfseries Limitations and Future Work:} Although we have achieved good results on image classification datasets, further exploration is needed on more complex tasks. In the future, we will focus on applying PPLL to more challenging tasks such as object detection, image segmentation, and image generation.

{
    \small
    \bibliographystyle{unsrt} 
    \bibliography{main}
}
\end{document}